\documentclass[]{llncs}

\usepackage{amssymb}
\usepackage{amsmath}
\usepackage{multicol}
\usepackage{pslatex}
\usepackage{multirow}
\usepackage{graphics}
\usepackage{graphicx}
\usepackage{url}
\usepackage{array}
\usepackage{color}

\def\ie{{\it i.e., }}   % %c'est à dire

\def\R{\mathbb R}  % %l'ensemble des reels= \R

\begin{document}

\title{Combination of Hidden Markov Random Field and  Conjugate Gradient  for Brain Image Segmentation}

\author{EL-Hachemi Guerrout\inst{1} \and Samy Ait-Aoudia\inst{1} \and Dominique Michelucci\inst{2} \and Ramdane Mahiou\inst{1}}

\institute{Ecole nationale Sup\'erieure en Informatique, Laboratoire LMCS, Oued-Smar, Algiers, Algeria,
\email{\{e\_guerrout, s\_ait\_aoudia, r\_mahiou\}@esi.dz}
\and Universit\'e de Bourgogne, Laboratoire LE2I, Dijon, France, 
\email{dominique.michelucci@u-bourgogne.fr}
}

\maketitle	

\begin{abstract}
Image segmentation is the process of partitioning the image into significant regions easier to analyze. Nowadays, segmentation has become a necessity in many practical medical imaging methods as locating tumors and diseases. Hidden Markov Random Field model is one of several techniques used in image segmentation. It provides an elegant way to model the segmentation process. This modeling leads to the minimization of an objective function. Conjugate Gradient algorithm (CG) is one of the best known optimization techniques. This paper proposes the use of the Conjugate Gradient algorithm (CG) for image segmentation, based on the Hidden Markov Random Field. Since derivatives are not available for this expression, finite differences are used in the CG algorithm to approximate the first derivative. The approach is evaluated using a number of publicly available images, where ground truth is known. The Dice Coefficient is used as an objective criterion to measure the quality of segmentation. The results show that the proposed CG approach compares favorably with other variants of Hidden Markov Random Field segmentation algorithms. 
\end{abstract}

\section{Introduction}
\label{sec:introduction}

\noindent Automatic segmentation of medical images becomes a crucial task due to  the huge amount of data produced by imaging devices. Many popular tools as FSL \cite{zhang2001segmentation} and Freesurfer \cite{fischl2012freesurfer} are dedicated to this aim.

There are several techniques to achieve the segmentation. We can broadly classify them into thresholding methods \cite{kumar2007skull,natarajan2012tumor,zhao2015automatic}, clustering methods \cite{pham2000current,wu1993optimal,chuang2006fuzzy}, edge detection methods \cite{perona1990scale,senthilkumaran2009edge,canny1986computational}, region-growing methods \cite{lin2012multispectral,roura2014marga}, watersheds methods \cite{benson2015brain,masoumi2012automatic}, model-based methods \cite{chan2001active,ho2002level,mcinerney1996deformable,wang2014segmentation} and Hidden Markov Random Field methods \cite{zhang2001segmentation,held1997markov,panjwani1995markov,guerrout2016hidden,guerrout2014hidden,ait2014medical,Guerrout:2016:HMR:3038884.3038886,guerrout2013medical,guerrout2013medicalconf}.

Threshold-based methods are the simplest ones that require only one pass through the pixels. They begin with the creation of an image histogram. After that, thresholds are used to separate the different image classes. For example, to segment an image into two classes, foreground and background, one threshold is necessary. The disadvantage of threshold-based techniques is the sensitivity to noise.

Region-based methods assemble neighboring pixels of the image in non-overlapping regions according to some homogeneity criterion. We distinguish two categories, region-growing methods and split-merge methods. They are effective when the neighboring pixels within one region have similar characteristics.

In model-based segmentation, a model is built for a specific anatomic structure by incorporating a prior information concerning shape, location and orientation. The presence of noise degrades the segmentation quality. This is why noise removal phase is generally an essential priority.

In classification methods, pixels are classified according to  some properties or criteria: gray level, texture or color.
 
Hidden Markov Random Field (HMRF) \cite{1984-geman-geman} provides an elegant way to model the segmentation problem. It is based on the MAP (Maximum A Posteriori) criterion \cite{wyatt2003map}. MAP estimation leads to  the minimization of an objective function \cite{szeliski2008comparative}. Therefore, optimization techniques are necessary to compute a solution. 
Conjugate Gradient Algorithm \cite{moller1993scaled,powell1977restart,shewchuk1994introduction} is one of the most popular optimization methods. 

This paper presents an automatic segmentation method based on the combination of Hidden Markov Field model and Conjugate Gradient algorithm. This method referred to as HMRF-CG, does not require preprocessing, feature extraction, training and learning. 
Brain MR image segmentation has attracted a particular attention in medical imaging. Thus, our tests rely on BrainWeb \footnote{\url{http://www.bic.mni.mcgill.ca/brainweb/}} \cite{cocosco1997brainweb} and IBSR\footnote{\url{https://www.nitrc.org/projects/ibsr}} images where the ground truth is known. 
Segmentation quality is evaluated using Dice Coefficient (DC) \cite{1945-dice} criterion. DC measures  how much the segmentation result is close to the ground truth. 
This paper is organized as follows. We begin by introducing the concept of Hidden Markov Field in section \ref{sec-hmrf}. A short section \ref{sec-cg} is devoted to the well known Conjugate Gradient algorithm. Section \ref{sec-exp-res} is devoted to the experimental results and section \ref{sec-con} concludes the paper.
\section{Hidden Markov Random Field (HMRF)}
\label{sec-hmrf}
\noindent Let $S=\{s_1,s_2,\dots,s_{M}\}$ be the sites or positions set. Both image to segment and segmented image are formed of $M$ sites. Each site $s \in S$ has a neighborhood set $ V_s (S) $. 
A neighborhood system $ V(S) $ has the following properties:
\begin{equation}
\left\{ 
\begin{array}{l}
\forall s \in S, s \notin V_s(S) \\
\forall \{s, t\} \in S, s \in V_t(S) \Leftrightarrow t \in V_s(S) 
\end{array}
\right.
\end{equation}
A $r$-order neighborhood system $ V^r(S) $ is defined by the following formula:
\begin{equation}
V_s^r(S) = \{ t \in S |\ \mbox{distance}(s,t)^2 \leq r^2 \wedge s \neq t \}
\end{equation}
\noindent where $\mbox{distance}(s,t)$ is the Euclidean distance between pixels $s$
and $t$. This distance depends only on the pixel position \ie it is not related to the pixel value. For volumetric data sets, as slices acquired by scanners, a 3D neighborhood system is used.

A clique $ c $ is a subset of $ S $ where all sites are neighbors to each other. 
For a non single-site clique, we have:
\begin{equation}
\forall \{s, t\} \in c, s \neq t \Rightarrow (t \in V_s(S) \wedge s \in V_t(S))
\end{equation}
A $p$-order clique noted $ C_p $ contains $ p $ sites i.e. $ p $ is the cardinal of the clique.

\begin{figure}[!h]
	\begin{center}
		\begin{tabular}{cc}
			\includegraphics[width=3cm,height=3cm]{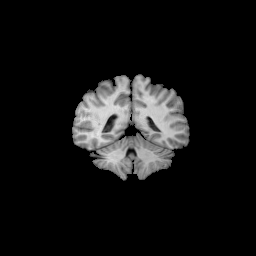}  &
			\includegraphics[width=3cm,height=3cm]{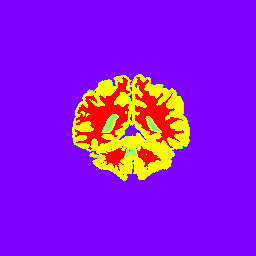} \\
			$y$: The image to segment & $x$: The segmented image\\
		\end{tabular}
		\caption{An example of segmentation with $K=4$.}
		\label{fig-obs-hidden}
	\end{center}
\end{figure}

Let $y=({y}_{1},{y}_{2},{\dots},{y}_{M })$ be the pixels values of the image to segment and $x=({x}_{1},{x}_{2},{\dots},{x}_{M })$ be the pixels classes of the segmented image.
$y_i$ and $x_i$ are respectively pixel value and class of the  site $s_i$.
The image to segment $y$ and the segmented image $x$ are seen respectively as a realization of Markov Random families $Y=({Y}_{1},{Y}_{2},{\dots},{Y}_{M })$ and $X=({X}_{1},{X}_{2},{\dots},{X}_{M })$.
The families of Random variables $ \{Y_s\}_{s \in S} $ and $ \{X_s\}_{s \in S} $ take their values respectively in the gray level space $ E_{y}=\{0,\ldots,255\} $ and the discrete space $E_x=\left\{1,{\dots},K\right\}$. 
$ K $ is the number of classes or homogeneous regions in the image. Configurations set of the image to segment $y$ and the segmented image are respectively $\Omega_y=E_{y}^M$ and $\Omega_x=E_{x}^M$. 
Figure \ref{fig-obs-hidden} shows an example of segmentation into four classes.

Segmentation of the image $y$ consists in looking for a realization $ x $ of $ X $. HMRF models this problem by maximizing the probability $P\left[X=x \;|\; {Y}=y\right]$.   

\begin{equation}
{x}^*=\operatorname*{arg}_{x \in \Omega_x}{\mathit{\max}}\left\{P[X=x\;|\;{Y}=y]\right\}
\end{equation}

From the Bayes rule, we get:
\begin{equation}\label{p-x-sha-y}
P\left[X=x\;|\;{Y}=y\right]=\frac{P\left[Y=y\;|\;{X}=x\right]\times
	P\left[X=x\right]}{P\left[Y=y\right]}
\end{equation}

Based on the conditional independence we have:

\begin{equation} \label{var-ind}
P\left[Y=y\;|\;{X}=x\right]=\prod
_{s{\in}S}{P[{Y}_{s}={y}_{s}\;|\;{{X}_{s}}={x}_{s}]}
\end{equation}

By the assumption that $P[{Y}_{s}={y}_{s}\;|\;{{X}_{s}}={x}_{s}]$ follows a normal distribution with mean ${\mu }_{x_s}$ and standard deviation ${\sigma }_{x_s}$, we will have:

\begin{equation} \label{loi-nor}
P\left[{Y}_{s}={y}_{s}\;|\;{{X}_{s}}=x_s\right]=\frac{1}{\sqrt{2\pi {\sigma
		}_{x_s}^{2}}}\mbox{exp}{\left(\frac{-{{({y}_{s}-{\mu }_{x_s})}^{2}}}{2{\sigma
	}_{x_s}^{2}}\right)}
\end{equation}
According to equation \ref{var-ind} and \ref{loi-nor}  we get: 

\begin{equation}
P\left[Y=y\;|\;{X}=x\right]=\prod_{s{\in}S}
\frac{1}{\sqrt{2\pi {\sigma
		}_{x_s}^{2}}}\mbox{exp}{\left(\frac{-{{({y}_{s}-{\mu }_{x_s})}^{2}}}{2{\sigma
	}_{x_s}^{2}}\right)}
\end{equation}

\begin{equation} \label{f1}
\Leftrightarrow P\left[Y=y\;|\;{X}=x\right]= {(2\pi)}^{\frac{-{M}}{2}}\ \mbox{exp}\ {\left(-\left( \sum_{ s \in S }{ \left[\ln (\sigma_{x_s}) + \frac{(y_s - \mu_{x_s})^2}{ 2 \sigma_{x_s}^2} \right] } \right)\right)}
\end{equation}

\noindent where $M$ is the image pixel number.

According to  Hammersley-Clifford theorem  \cite{1971-hammersley-clifford} which establishes the equivalence between Markov field and Gibbs, we get:
\begin{equation} \label{p-x}
P[X=x]=\frac{\mbox{exp} \left({\frac{-U(x)}{T}}\right)}{\sum _{\xi \in \Omega_x }{\mbox{exp}{\left(\frac{-U(\xi)}{T}\right)}}}
\end{equation}
\noindent where $ T $ is a control parameter called temperature.

The energy $U(x)$ is defined by Potts model \cite{swendsen1987nonuniversal} as follows:

\begin{equation}{\label{plot}}
U(x)=\beta \sum _{{c}_{2}=\{s,t\}}{(1-2\delta
	({x}_{s},{x}_{t}))}
\end{equation}

\noindent where $\beta$ is a constant and $\delta $ is the Kronecker's delta:
\begin{equation}
\delta 
(a,b)=\left\{
\begin{array}{lll}
1 & \mbox {if} & a=b\\
0 & \mbox {if} & a{\neq}b
\end{array}
\right.
\end{equation}

${P\left[Y=y\right]}$ is a constant, so pose: 

\begin{equation}\label{f3}
P[Y=y] =  \mbox{C}
\end{equation}

By replacing the equations (\ref{f1}), (\ref{p-x}) and (\ref{f3}) in the equation (\ref{p-x-sha-y}), we will have:

\begin{equation}
\left\{ 
\begin{array}{l}
P[X=x|Y=y] = \mbox{A} \exp\left({-\Psi(x,y)}\right)\\
\Psi (x,y) = 
\sum_{ s \in S }{ \left[\ln (\sigma_{x_s}) + \frac{(y_s - \mu_{x_s})^2}{ 2 \sigma_{x_s}^2} \right] } 
+ \frac{\beta}{T} \sum_{c_2 = \{s,t\}}{ ( 1 - 2 \delta(x_s,x_t) ) }\\
\mbox{A is a positive constant} 
\end{array}
\right.
\end{equation}
\noindent where $ T $ is a control parameter called temperature,   
$ \delta $ is a Kronecker's delta and $ \mu_{x_s} $, $ \sigma_{x_s} $ are respectively the mean and standard deviation of the class $x_s$.
When  $\beta >0$, the most likely segmentation corresponds to the constitution of large homogeneous regions. The size of these regions is controlled by the $\beta $ value.

Maximizing the probability $ P[X=x\;|\;Y=y] $ is equivalent to minimizing the function $ \Psi(x,y) $.
\begin{equation}
{x}^*=\operatorname*{arg}_{x \in \Omega_x}{\mathit{\min}}\left\{\Psi(x,y)\right\}
\end{equation}

The computation of the exact segmentation ${x}^*$  is impossible \cite{1984-geman-geman}. Therefore optimization techniques are necessary to compute an approximate solution $\hat{{x}}$. 

Let $\mu=(\mu_1,\dots,\mu_j,\dots,\mu_K)$ be the means and $\sigma=(\sigma_1,\dots,\sigma_j,\dots,\sigma_K)$ be the standard deviations of $K$ classes in the segmented image $x=(x_1,\dots,x_s,\dots,x_M)$ \ie
\begin{equation}{\label{sj}}
\begin{array}{l}
\begin{cases}
\mu_j={\frac{1}{|S_j|} \sum_{s \in S_j} y_s}\\
\sigma_j=\sqrt{\frac{1}{|S_j|} \sum_{s \in S_j} (y_s-\mu_j)^2}\\
S_j=\{s\ |\ x_s=j\}
\end{cases}
\end{array}
\end{equation}

Our way to minimize the function $\Psi(x,y) $ is to minimize instead 
the function $\Psi(\mu)$. We can always compute $x$ through $\mu$ by classifying $y_s$ into the nearest mean $\mu_j$ \ie $x_s=j$ if the nearest mean to $y_s$ is $\mu_j$. Thus instead of looking for $x^*$, we look for $\mu^*$. The configuration set of $\mu$ is $\Omega_{\mu}=[0\dots255]^K$.
\begin{equation}
\begin{array}{l}
\begin{cases}
{\mu}^*=\operatorname*{arg}_{\mu \in \Omega_{\mu}}{\mathit{\min}}\left\{\Psi(\mu)\right\}\\
\Psi(\mu) = \sum_{ j = 1 }^{K} \sum\limits_{s \in S_j }{ [\ln (\sigma_j ) + \frac{ (y_s - \mu_j)^2 }{ 2 \sigma_j^2 }] }  +  \frac{\beta}{T} \sum_{c_2 = \{s,t\}}{ ( 1 - 2 \delta(x_s,x_t) ) } 
\end{cases}
\end{array}
\end{equation}

\noindent where $S_j$, $\mu_j$ and $\sigma_j$ are defined in the equation (\ref{sj}).

To apply unconstrained optimization techniques,  we redefine the function $\Psi(\mu)$ for $\mu \in \R^K$ instead of $\mu \in \Omega_{\mu}$. Therefore, the new function  $\Psi(\mu)$ becomes as follows: 

\begin{equation} \label{eq-eq}
\begin{array}{l}
\Psi(\mu) =
\begin{cases}
\sum_{ j = 1 }^{K} \sum\limits_{s \in S_j }{ [\ln (\sigma_j ) + \frac{ (y_s - \mu_j)^2 }{ 2 \sigma_j^2 }] }  +  \frac{\beta}{T} \sum_{c_2 = \{s,t\}}{ ( 1 - 2 \delta(x_s,x_t) ) }   \ \mbox {if} \ \mu\in \Omega_{\mu}\\
+\infty \	\  \mbox {otherwise}
\end{cases}
\end{array}
\end{equation}

\section{The Conjugate Gradient (CG) Algorithm} \label{sec-cg}

\noindent  In practice, the application is implemented in the cross-platform Qt creator (C++) under Linux system. We have used the GNU Scientific Library implementation of Polak-Ribi{\`e}re  Conjugate Gradient method \cite{polak1969note,grippo1997globally} (gsl$\_$multimin$\_$fdfminimizer$\_$conjugate$\_$pr).

To use Conjugate Gradient Algorithm, we need the first derivative $\Psi^{'}(\mu))=(d_1,\dots,d_i,\dots,d_n)$. 
Since no mathematical expression is available,
it is approximated with finite differences \cite{eberly2003derivative} as follows: 
\begin{itemize}
	\item A forward difference approximation is
	\begin{equation}
	d_i=\frac{\Psi(\mu_1,\dots,\mu_i+\varepsilon,\dots,\mu_n)-\Psi(\mu_1,\dots,\mu_i,\dots,\mu_n)}{\varepsilon} 
	\end{equation}
	
	\item A backward difference approximation is
	\begin{equation}
	d_i=\frac{\Psi(\mu_1,\dots,\mu_i,\dots,\mu_n)-\Psi(\mu_1,\dots,\mu_i-\varepsilon,\dots,\mu_n)}{\varepsilon} 
	\end{equation}
	
	\item A centered difference approximation is
	\begin{equation}
	d_i=\frac{\Psi(\mu_1,\dots,\mu_i+\varepsilon,\dots,\mu_n)-\Psi(\mu_1,\dots,\mu_i-\varepsilon,\dots,\mu_n)}{2\varepsilon} 
	\end{equation}
	
\end{itemize}

In our tests, we have used a centered difference approximation to compute the first derivative.
The good approximation of the first derivative relies  on the choice of the value of the parameter $\epsilon$. Through the tests conducted, we have selected $0.01$ as the best value. 

\section{Experimental Results}
\label{sec-exp-res}

In this section, we begin by showing the effectiveness of HMRF-CG method. To this end, we will make a comparison with some methods (used in the field) that are: MRF-Classical \cite{2012-yousefi-al}, MRF-ACO-Gossiping \cite{2012-yousefi-al} and MRF-ACO \cite{2003-salima-mohamed}. Secondly, we will show the robustness of HMRF-CG method against noise, by doing a comparison with LGMM method (Local Gaussian Mixture Model)\cite{2013-liu-zhang}. 
The implementation of LGMM  is built upon the segmentation method \cite{ashburner2005unified} of 
SPM 8 (Statistical Parametric Mapping\footnote{http://www.fil.ion.ucl.ac.uk/spm/software/spm8/}), which is a well known software for MRI
analysis. As reported by \cite{2013-liu-zhang}, LGMM has better results than SPM 8. 

To perform a fair and meaningful comparison, we have used a metric known as Dice Coefficient \cite{1945-dice}. Morey et al. \cite{morey2009comparison} used interchangeably Dice coefficient and Percentage volume overlap. This metric is usable only when the ground truth segmentation is known (see section \ref{DC}). The image sets and related parameters are described in section \ref{imaset}. Finally, section \ref{R} is devoted to the yielded results.

\subsection{Dice Coefficient metric} \label{DC} 
\noindent Dice Coefficient (DC) measures how much the result is close to the ground truth. Let the resulting class be $\hat{A}$ and its ground truth be $A^*$. Dice Coefficient is given by the following formula:
\begin{equation}
\mathit{DC} 	= 	\frac{ 2 |\hat{A} \cap A^*| }{\ |\hat{A} \cup A^*| } 
=	\frac{ 2 \mbox{TP}}{ 2 \mbox{TP} +  \mbox{FP} + \mbox{FN} }
\end{equation}
where TP stands for true positive, FP for false positive and FN for false negative.
DC equals 1 in the best case \ie $\hat{A}$ and  $A^*$ are identical and it equals 0 in the worst case \ie  there is an empty intersection between  $\hat{A}$ and  $A^*$.

\begin{figure}[!h]
	\begin{center}
		\includegraphics[scale=1]{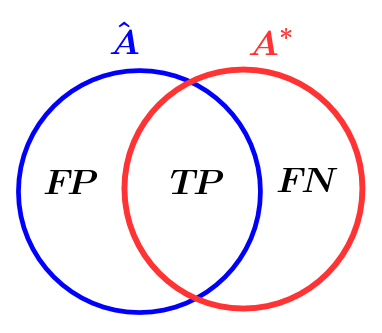} % % %
		\caption{TP, FP and FN.}
		\label{fig-KI}
	\end{center}
\end{figure}

\subsection{The image sets and related parameters}\label{imaset}

To evaluate the quality of segmentation, we use four volumetric (3D) MR images, one obtained from IBSR (real image) and the others from BrainWeb (simulated images). Three components were considered: GM (Grey Matter), WM (White Matter) and CSF (Cerebro Spinal Fluid).

IBSR image dimension is $256\times256\times63$, with voxel=$1\times3\times1$mm and T1-weighted modality. 
The three BrainWeb images dimensions are $181\times217\times181$, with voxels=$1\times1\times1$mm and T1-weighted modality. We tested three levels of noise $0\%$, $3\%$ and $5\%$ with different intensity non-uniformity $0\%$ and $20\%$.

In this paper we have retained a subset of slices, which are cited in \cite{2012-yousefi-al}. The IBSR slices retained are: 1-24/18, 1-24/20, 1-24/24, 1-24/26, 1-24/30, 1-24/32 and 1-
24/34. The BrainWeb slices retained are: 85, 88, 90, 95, 97, 100, 104, 106, 110, 121 and 130. 

Table \ref{tab-RP} defines some parameters necessary to execute HMRF-CG method.
\begin{table}[!h]
	\footnotesize
	\setlength{\tabcolsep}{0.75pt}
	\setlength{\extrarowheight}{0.75pt}
	\centering
	\caption{\label{tab-RP} Related parameters to the images used in our tests.}
	\begin{tabular}{|l|c|c|c|}
		\hline
	Image	& The constant $\beta$ & The temperature $T$ 		& The initial point $\mu^0 $	\\
		\hline
	IBSR 		& \multirow{4}{*}{1} & 10  	 	& (1, 5, 140, 190) \\
	\cline{1-1}\cline{3-4}
	BrainWeb1	&  & 10  	 	& \multirow{3}{*}{(1, 45, 110, 150)} \\
	\cline{1-1}\cline{3-3}
	BrainWeb2	&  & 4  	 	&  \\
	\cline{1-1}\cline{3-3}
	BrainWeb3	&  & 1  	 	&  \\
	\hline
	\end{tabular}
\end{table}

\subsection{Results}\label{R}

Table \ref{tab-dc} shows the mean DC values using IBSR image. The parameters used by HMRF-CG are described in Table \ref{tab-RP}. The parameters used by the other methods are given in \cite{2012-yousefi-al,2003-salima-mohamed}.

\begin{table}[h!]
	\footnotesize
	\setlength{\tabcolsep}{0.75pt}
	\setlength{\extrarowheight}{0.75pt}
	\centering
	\caption{\label{tab-dc} Mean DC values (the best results are given in bold type).}
	\begin{tabular}{|l|c|c|c|c|}
		\hline
		\multirow{2}{*}{Methods}  			& \multicolumn{4}{c|}{Dice Coefficient} \\
		\cline{2-5}
		& GM & WM 		& CSF 		& Mean \\
		\hline
		Classical-MRF 		& 0.771
		& 0.828  	 	& 0.253 	& 0.617 \\
		\hline
		MRF-ACO 			& 0.778 	& 0.827 	& 0.263
		& 0.623 \\
		\hline
		MRF-ACO-Gossiping 	& 0.778 	& 0.827 	& 0.262 	& 0.623 \\
		\hline
		HMRF-CG & \textbf{0.859} 	& \textbf{0.855} 	& \textbf{0.381} 	& \textbf{0.698} \\
		\hline
	\end{tabular}
\end{table}

Table \ref{tab-dc2} shows the mean DC values using BrainWeb images. The parameters used by HMRF-CG are described in Table \ref{tab-RP}. The parameters used by the LGMM method are given in LGMM \cite{2013-liu-zhang}. The column (N,I) gives noise and intensity non-uniformity.

\begin{table}
	\footnotesize
	\setlength{\tabcolsep}{0.75pt}
	\setlength{\extrarowheight}{0.75pt}%
	\centering
	\caption{\label{tab-dc2} Mean DC values (the best results are in bold type).}
	\begin{tabular}{|c|c|c|c|c|c|c|c|c|c|}
		\hline
		\multirow{3}{*}{Image} &
		\multirow{3}{*}{(N,I)} &
		 \multicolumn{4}{c|}{HMRF-CG} & \multicolumn{4}{c|}{LGMM}\\
		\cline{3-10}
		& &\multicolumn{4}{c|}{Dice Coefficient} & 
		\multicolumn{4}{c|}{Dice Coefficient}\\
		\cline{3-10}
		 & & GM & WM & CSF & Mean & GM & WM & CSF & Mean\\
		\hline
		BrainWeb1&(0\%,0\%)  &\textbf{0.970} & \textbf{0.990} & \textbf{0.961} & \textbf{0.974}
		& 0.697 & 0.667 & 0.751 & 0.705
		\\
		\hline
		
		BrainWeb2&(3\%,20\%)  & \textbf{0.940} & \textbf{0.965} & \textbf{0.940} & \textbf{0.949}
		& 0.905 & 0.940 & 0.897 & 0.914 
		\\
		\hline
		
		BrainWeb3& (5\%,20\%)  & \textbf{0.918} & \textbf{0.952} & \textbf{0.924} &\textbf{0.931} &
		0.912 & 0.951 & 0.893 & 0.918 
		\\
		\hline
	\end{tabular}
\end{table}

\begin{figure}[!h]
	\renewcommand{\arraystretch}{1}
	\begin{center}
		\setlength{\tabcolsep}{2pt}
		\setlength{\extrarowheight}{2pt}%
		\footnotesize
		\newlength{\iw}
		\newlength{\ih}
		\renewcommand{\angle}{0}
		\setlength{\iw}{3cm}
		\setlength{\ih}{3cm}
		
		\begin{tabular}{|c|c|}
			\hline
			\includegraphics[width=\iw,height=\ih,angle=\angle]{img/18_.png}  &
			\includegraphics[width=\iw,height=\ih,angle=\angle]{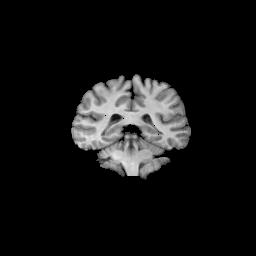} 
			\\ 
			\hline
			IBSR 1-24/18 & IBSR 1-24/20 
			\\
			\hline
			\includegraphics[width=\iw,height=\ih,angle=\angle]{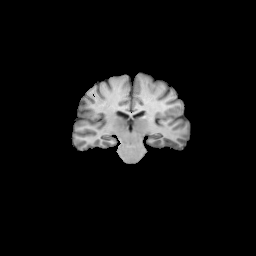}  &
			\includegraphics[width=\iw,height=\ih,angle=\angle]{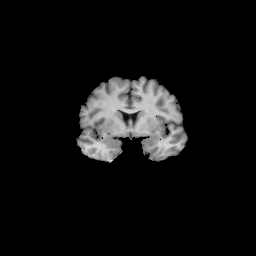} 
			\\
			\hline
			IBSR 1-24/24 & IBSR 1-24/32 
			\\
			\hline
			
		\end{tabular}
		
		\caption{A sample of slices to segment from IBSR image.}
		\label{fig-ibsr}
	\end{center}
\end{figure}

\begin{figure}[!h]
	\renewcommand{\arraystretch}{1}
	\begin{center}
		\setlength{\tabcolsep}{2pt}
		\setlength{\extrarowheight}{2pt}%
		\footnotesize
		%\newlength{\iw}
		%\newlength{\ih}
		\renewcommand{\angle}{0}
		\setlength{\iw}{3cm}
		\setlength{\ih}{3cm}
		
		\begin{tabular}{|c|c|}
			\hline
			\includegraphics[width=\iw,height=\ih,angle=\angle]{img/18.png}  &
			\includegraphics[width=\iw,height=\ih,angle=\angle]{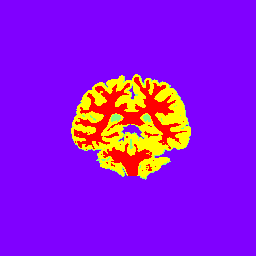} 
			\\ 
			\hline
			IBSR 1-24/18 & IBSR 1-24/20 
			\\
			\hline
			\includegraphics[width=\iw,height=\ih,angle=\angle]{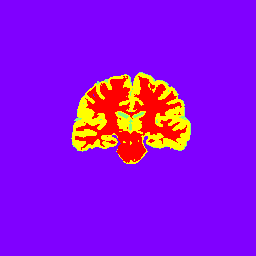}  &
			\includegraphics[width=\iw,height=\ih,angle=\angle]{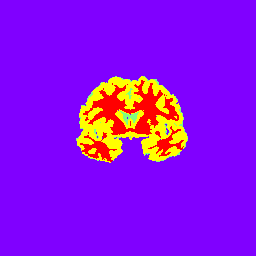} 
			\\
			\hline
			IBSR 1-24/24 & IBSR 1-24/32 
			\\
			\hline
			
		\end{tabular}
		
		\caption{A sample of segmented slices using HMRF-CG.}
		\label{fig-ibsr2}
	\end{center}
\end{figure}

\begin{figure}[!h]
	\renewcommand{\arraystretch}{1}
	\begin{center}
		\setlength{\tabcolsep}{2pt}
		\setlength{\extrarowheight}{2pt}%
		\footnotesize
		%\newlength{\iw}
		%\newlength{\ih}
		\renewcommand{\angle}{0}
		\setlength{\iw}{3cm}
		\setlength{\ih}{3cm}
		
		\begin{tabular}{|c|c|}
			\hline
			\includegraphics[width=\iw,height=\ih,angle=\angle]{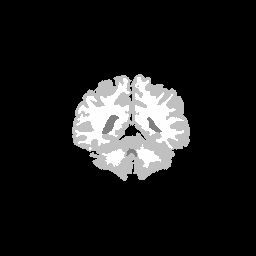}  &
			\includegraphics[width=\iw,height=\ih,angle=\angle]{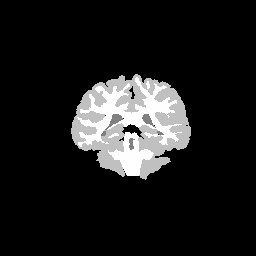} 
			\\ 
			\hline
			IBSR 1-24/18 & IBSR 1-24/20 
			\\
			\hline
			\includegraphics[width=\iw,height=\ih,angle=\angle]{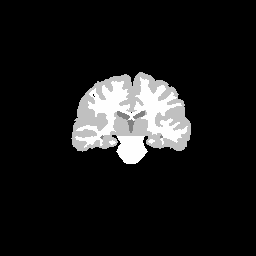}  &
			\includegraphics[width=\iw,height=\ih,angle=\angle]{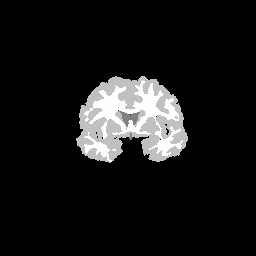} 
			\\
			\hline
			IBSR 1-24/24 & IBSR 1-24/32 
			\\
			\hline
			
		\end{tabular}
		
		\caption{A sample of ground truths images from IBSR image.}
		\label{fig-ibsr3}
	\end{center}
\end{figure}

Figure \ref{fig-ibsr}, Figure \ref{fig-ibsr2} and Figure \ref{fig-ibsr3} show respectively a sample of slices to  segment obtained from IBSR image, a segmented slice using HMRF-CG method and a ground truths slice.

\begin{figure}[!h]
	\begin{center}
		\setlength{\tabcolsep}{2pt}
		\setlength{\extrarowheight}{2pt}%
		\footnotesize
		
		\renewcommand{\angle}{0}
		\setlength{\iw}{3cm}
		\setlength{\ih}{4cm}
		
		\begin{tabular}{|>{\centering\arraybackslash}m{2cm}|>{\centering\arraybackslash}m{2cm}|>{\centering\arraybackslash}m{3cm}|>{\centering\arraybackslash}m{3cm}|}
			\hline
			Image &(N,I) & Slice to segment &  Segmented slice\\
			\hline
			
			BrainWeb1&(0\%,0\%) &
			\includegraphics[width=\iw,height=\ih,angle=\angle]{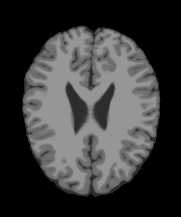}  &
			\includegraphics[width=\iw,height=\ih,angle=\angle]{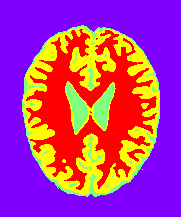} \\
			\hline
			BrainWeb2&(3\%,20\%) &
			\includegraphics[width=\iw,height=\ih,angle=\angle]{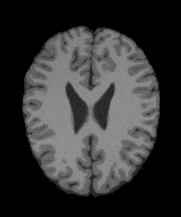}  &
			\includegraphics[width=\iw,height=\ih,angle=\angle]{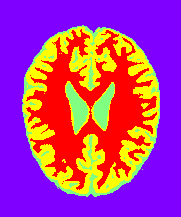} \\
			\hline
			BrainWeb3&(5\%,20\%) &
			\includegraphics[width=\iw,height=\ih,angle=\angle]{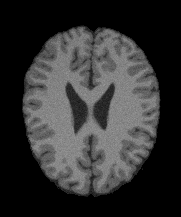}  &
			\includegraphics[width=\iw,height=\ih,angle=\angle]{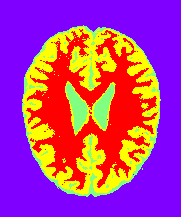} \\
			\hline
			
		\end{tabular}
		
		\caption{The slices number \#97 with different noise and intensity non-uniformity from BrainWeb images and their segmentation using HMRF-CG.}
		\label{fig-brainweb}
	\end{center}
\end{figure}

Figure \ref{fig-brainweb} shows a sample of slices to segment from BrainWeb images and segmented slices using HMRF-CG. The column (N,I) gives noise and intensity non-uniformity. 

\section{Discussion and Conclusion}
\label{sec-con}
\noindent In this paper, we have described a method which combines Hidden Markov Random Field (HMRF) and Conjugate Gradient (GC). The tests have been carried out on samples obtained from IBSR and BrainWeb images, the most commonly used images in the field. For a fair and meaningful comparison of methods, the segmentation quality is measured using the Dice Coefficient metric. The results depend on the choice of parameters. This very sensitive task has been conducted by performing numerous tests. From the results obtained, the HMRF-GC method outperforms the methods tested that are: LGMM, Classical MRF, MRF-ACO-Gossiping and MRF-ACO. 
Tests permit to find  good parameters for HMRF-CG to achieve good segmentation results. To further improve performances a preprocessing step can be added to reduce noise and inhomogeneity using appropriate filters.

\section*{Acknowledgment}

\noindent  We would like to thank BRAINWEB and IBSR communities for allowing us to use their images in the evaluation of brain segmentation.

\bibliographystyle{splncs03}
\bibliography{strings}
\end{document}